\documentclass[runningheads]{llncs}
\usepackage{graphicx}
\usepackage{comment}
\usepackage{amsmath,amssymb} % define this before the line numbering.
\usepackage{color}
\usepackage{multirow}
\usepackage{booktabs}

\usepackage{array}

\newcolumntype{Y}{>{\centering\arraybackslash}m{1cm}}
\usepackage{footmisc}
\usepackage{siunitx,booktabs,etoolbox}
\usepackage[colorlinks=true,citecolor=blue,urlcolor=black]{hyperref}
\usepackage[misc]{ifsym}
\usepackage{algorithm,algcompatible,amsmath}
\providecommand{\mb}[1]{\mathbf{#1}}

\providecommand{\mcl}[1]{\mathcal{#1}} 

\providecommand{\mbx}{\mb{x}}
\providecommand{\mby}{\mb{y}}

\providecommand{\cjh}{c^{\hat{j}}}
\providecommand{\cj}{c^j}

\makeatletter
\newcommand{\printfnsymbol}[1]{%
  \textsuperscript{\@fnsymbol{#1}}%
}
\makeatother

\newcommand\blfootnote[1]{%
  \begingroup
  \renewcommand\thefootnote{}\footnote{#1}%
  \addtocounter{footnote}{-1}%
  \endgroup
}
\usepackage{xcolor}

\begin{document}
% \linenumbers
\pagestyle{headings}
\mainmatter
\def\ECCVSubNumber{6977}  % Insert your submission number here

\title{Self-supervision with Superpixels: \\
Training Few-shot Medical Image Segmentation without Annotation} %
\begin{comment}
\titlerunning{ECCV-20 submission ID \ECCVSubNumber} 
\authorrunning{ECCV-20 submission ID \ECCVSubNumber} 
\author{Anonymous ECCV submission}
\institute{Paper ID \ECCVSubNumber}
\end{comment}
%******************

%\begin{comment}
\titlerunning{Self-supervised Learning for Few-shot Medical Image Segmentation}
\author{Cheng Ouyang\inst{1}$^{\textrm{(\Letter)}}$  \and
Carlo Biffi\inst{1}\thanks{Equal contribution.} \and
Chen Chen\inst{1}\printfnsymbol{1} \and
Turkay Kart\inst{1}\printfnsymbol{1} \and \\
Huaqi Qiu\inst{1}\and
Daniel Rueckert\inst{1}
}
\authorrunning{Ouyang et al.}
\institute{BioMedIA Group, Department of Computing, Imperial College London, UK 
\email{c.ouyang@imperial.ac.uk}}
%\end{comment}
%******************
\maketitle
\begin{abstract}
Few-shot semantic segmentation (FSS) has great potential for medical imaging applications. Most of the existing FSS techniques require abundant annotated semantic classes for training. However, these methods may not be applicable for medical images due to the lack of annotations. To address this problem we make several contributions: (1) A novel self-supervised FSS framework for medical images in order to eliminate the requirement for annotations during training. Additionally, superpixel-based pseudo-labels are generated to provide supervision; (2) An adaptive local prototype pooling module plugged into prototypical networks, to solve the common challenging foreground-background imbalance problem in medical image segmentation; (3) We demonstrate the general applicability of the proposed approach for medical images using three different tasks: abdominal organ segmentation for CT and MRI, as well as cardiac segmentation for MRI. Our results show that, for  medical image segmentation, the proposed method outperforms conventional FSS methods which require manual annotations for training. 
%\keywords{few-shot segmentation; self-supervised learning; medical images}
\end{abstract}
\blfootnote{Code is available at \url{https://github.com/cheng-01037/Self-supervised-Fewshot-Medical-Image-Segmentation}}

\section{Introduction}
Automated medical image segmentation is a key step for a vast number of clinical procedures and medical imaging studies, including disease diagnosis and follow-up \cite{pham2000current,sharma2010automated,zhang2011multimodal}, treatment planning \cite{el2007concurrent,zaidi2010pet} and population studies \cite{de2001prevalence,petersen2013imaging}.
Fully supervised deep learning based segmentation models can achieve good results when trained on abundant labeled data. However, the training of these networks in medical imaging is often impractical due to the following two reasons: there is often a lack of sufficiently large amount of expert-annotated data for training due the considerable clinical expertise, cost and time associated with annotation; This problem is further exacerbated by differences in image acquisition procedures across medical devices and hospitals, often resulting in datasets containing few manually labeled images; 
Moreover, the number of possible segmentation targets (different anatomical structures, different types of lesions, etc.) are countless. It is impractical to cover every single unseen class by training a new, specific model.

As a potential solution to these two challenges, few-shot learning has been proposed \cite{snell2017prototypical,sung2018learning,garcia2017few,vinyals2016matching,fei2006one,lake2011one}. During \textit{inference}, a few-shot learning model distills a discriminative representation of an unseen class from only a few labeled examples (usually denoted as \textit{support}) to make predictions for unlabeled examples (usually denoted as \textit{query}) without the need for re-training the model. If applying few-shot learning to medical images, segmenting a rare or novel lesion can be potentially efficiently achieved using only a few labeled examples.

\begin{figure}[!t]
\centering
\includegraphics[width=11cm]{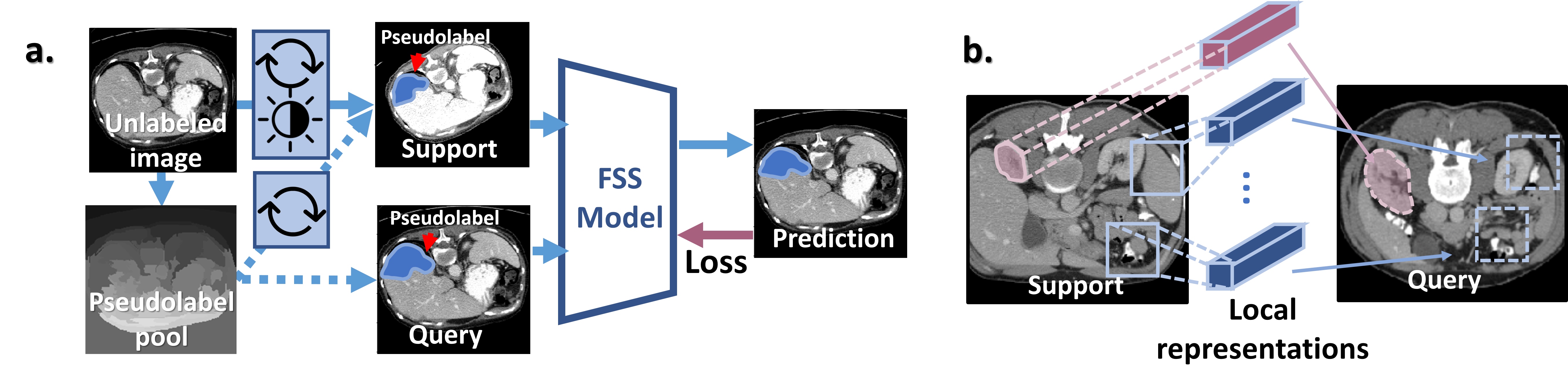}
\caption{ (a). Proposed superpixel-based self-supervised learning. For each unlabeled image, pseudolabels are generated on superpixels. In each iteration during training, a randomly selected pseudolabel and the original image serve as the candidate for both support and query. Then, random transforms (marked in blue boxes) are applied between the support and the query. The self-supervision task is designed as segmenting the pseudolabel on the query with reference to the support, despite the transforms applied in between. (b). The proposed ALPNet solves the class-imbalance-induced ambiguity problem by adaptively extracting multiple local representations of the large background class (in blue). Each of them only represents a local region of background.}
\label{fig: intro}
\end{figure}

However, \textit{training} an existing few-shot semantic segmentation (FSS) model for medical imaging has not had much success in the past, as most of FSS methods rely on a large training dataset with many annotated training classes to avoid overfitting \cite{rakelly2018conditional,shaban2017one,dong2018few,siam2019amp,wang2019panet,zhang2018sg,rakelly2018few,zhang2019pyramid,siam2019amp,siam2020weakly,tian2019differentiable,hu2019silco,hendryx2019meta}. In order to bypass this unmet need of annotation, we propose to train an FSS model on unlabeled images instead via self-supervised learning, an unsupervised technique that learns generalizable image representations by solving a carefully designed task \cite{lieb2005adaptive,doersch2015unsupervised,dosovitskiy2014discriminative,larsson2016learning,noroozi2016unsupervised,gidaris2019boosting,pathak2016context,zhang2016colorful}. 
Another challenge for a lot of state-of-the-art FSS network architectures is the loss of local information within a spatially variant class in their learned representations. This problem is in particular magnified in medical images since extreme foreground-background imbalance commonly exists in medical images. 
As shown in Fig. \ref{fig: intro} (b)., the background class is large and spatially inhomogeneous whereas the foreground class (in purple) is small and homogeneous.  Under this scenario, an ambiguity in prediction on foreground-background boundary might happen if the distinct appearance information of different local regions (or saying, parts) in the background is unreasonably averaged out. Unfortunately, this loss of intra-class local information exists in a lot of recent works, where each class is spatially averaged into a 1-D representation prototype  \cite{dong2018few,wang2019panet,zhang2018sg,liu2020prototype} or weight vectors of a linear classifier \cite{siam2019amp}. In adjust to this problem, we instead encourage the network to preserve intra-class local information, by extracting an ensemble of local representations for each class. 

In order to break the deadlock of training data scarcity and to boost segmentation accuracy, we propose SSL-ALPNet, a self-supervised few-shot semantic segmentation framework for medical imaging. The proposed framework exploits \textit{superpixel-based self-supervised learning} (SSL), using superpixels for eliminating the need for manual annotations, and an \textit{adaptive local prototype pooling} enpowered prototypical network (ALPNet), improving segmentation accuracy by preserving local information in learned representations. As shown in Fig. \ref{fig: intro} (a), to ensure image representations learned through self-supervision are well-generalizable to real semantic classes, we generate pseudo-semantic labels using superpixels, which are compact building blocks for semantic objects \cite{ren2003learning,stutz2018superpixels,achanta2010slic}. In addition, to improve the discriminative ability of learned image representations, we formulate the self-supervision task as one-superpixel-against-the-rest segmentation. Moreover, to enforce invariance in representations %against intensity and shape variants 
between support and query, which is crucial for few-shot segmentation in real-world, we synthesis variants in shape and intensity by applying random geometric and intesity transforms between support and query. 
In our experiments, we observed that by purely training with SSL, our network outperforms those trained with manual annotated classes by considerable margins. Besides, as shown in Fig. \ref{fig: intro}, 
to boosts segmentation accuracy, we designed adaptive local prototype module (ALP) for preserving local information of each class in their prototypical representations. This is achieved by extracting an ensemble of local representation prototypes, each focuses on a different region. Of note, the number of prototypes are allocated adaptively by the network based on the spatial size of each class. By this mean, ALP alleviates ambiguity in segmentation caused by insufficient local information. 

Overall, the proposed SSL-ALPNet framework has the following major advantages: Firstly, compared with current state-of-the-art few-shot segmentation methods which in general rely on a large number of annotated classes for training, the proposed method eliminates the need for annotated training data instead. By completely detaching representation extraction from manual labeling, the proposed method potentially expands the application of FSS in annotation-scarce medical images. In addition, unlike most of self-supervised learning methods for segmentation where fine-tuning on labeled data is still required before testing \cite{doersch2015unsupervised,dosovitskiy2014discriminative,larsson2016learning,noroozi2016unsupervised,pathak2016context,zhou2019models}, the proposed method requires no fine-tuning after SSL.
Moreover, compared to some of novel modules \cite{vaswani2017attention,schlichtkrull2018modeling,kipf2016semi} used in FSS where slight performance gain are at the cost of heavy computations, the proposed ALP is simple and efficient in contrast to its significant performance boosting. No trainable parameters are contained in ALP. 

Our contributions are summarized as follows:
\begin{itemize}
    \item We propose SSL-ALPNet, the first work that explores self-supervised learning for few-shot medical image segmentation, to the best of our knowledge. It outperforms peer FSS methods, which usually require training with manual annotations, by merely training on unlabeled images.
    \item We propose adaptive local prototype pooling, a local representation computation module that significantly boosts performance of the state-of-the-art prototypical networks on medical images. 
    \item We for the first time evaluated FSS on different imaging modalities, segmentation classes and with the presence of patient pathologies. The established evaluation strategy not only highlights wide applicability of our work, but also facilitates future works that seek to evaluate FSS in a more realistic scenario.
\end{itemize}

\section{Related Work}

\subsection{Few-shot semantic segmentation}
\label{subsec: fss_review}
Recent work by \cite{gidaris2019boosting} firstly introduces self-supervised learning into few-shot image classification. However, few-shot segmentation is often more challenging: dense prediction needs to be performed at a pixel level. 
To fully exploit information in limited support data, most of popular FSS methods directly inject support to the network as guiding signals \cite{rakelly2018few,zhang2019pyramid,zhang2019canet,roy2020squeeze}, or construct discriminative representations from support as reference to segment query \cite{shaban2017one,dong2018few,siam2019amp,zhang2018sg,wang2019panet}. 
The pioneering work \cite{shaban2017one} learns to generate classifier weights from support; \cite{siam2019amp} extends weights generation to multi-scale. \cite{rakelly2018conditional} instead directly uses support to condition segmentation on query by fusing their feature maps. Exploiting network components such as attention modules \cite{vaswani2017attention,hu2019attention} and graph networks \cite{kipf2016semi,schlichtkrull2018modeling}, recent works boost segmentation accuracy \cite{zhang2019pyramid} and enable FSS with coarse-level supervisions \cite{zhang2019canet,siam2020weakly,hu2019silco}. % by introducing suitable inductive bias through network design. 
Exploiting learning-based optimization, \cite{tian2019differentiable,hendryx2019meta} combine meta-learning with FSS. However, almost all of these methods assume abundant annotated (including weakly annotated) training data to be available, making them difficult to translate to segmentation scenarios in medical imaging.  

One main stream of FSS called \textit{prototypical networks} focuses on exploiting \textit{representation prototypes} of semantic classes extracted from the support. These prototypes are utilized to make similarity-based prediction \cite{snell2017prototypical,wang2019panet,liu2020prototype} on query, or to tune representations of query \cite{dong2018few}. Recently, prototypical alignment network (PANet) \cite{wang2019panet} has achieved state-of-the-art performance on natural images. This is achieved simply with a generic convolutional network and an alignment regularization. However, these works aim to improve performance on training-classes-abundant natural images. Their methodologies focus on network design. Our work, by contrast, focuses on utilizing unlabeled medical image for training by exploiting innovative training strategies and pesudolabels. Nevertheless, since PANet is one of state-of-the-art and is conceptually simple, we take this method as our baseline to highlight our self-supervised learning as a generic training strategy.

In medical imaging, most of recent works on few-shot segmentation only focus on training with less data \cite{zhao2019data,mondal2018few,ouyang2019data,yu2020foal,chen2020realistic}. These methods usually still require re-training before applying to unseen classes, and therefore they are out-of-scope in our discussion. Without retraining on unseen classes, the SE-Net \cite{roy2020squeeze} introduces squeeze and excite blocks \cite{hu2018squeeze} to \cite{rakelly2018conditional}. To the best of our knowledge, it is the first FSS model specially designed for medical images, with which we compared our method in experiments. 

\subsection{Self-supervised learning in semantic segmentation}
A series of self-supervision tasks have been proposed for semantic segmentation. Most of these works focus on intuitive handcrafted supervision tasks including spatial transform prediction \cite{doersch2017multi}, image impainting \cite{pathak2016context}, patch reordering \cite{doersch2015unsupervised}, image colorization \cite{zhang2016colorful}, difference detection \cite{shimoda2019self}, motion interpolation \cite{zhan2019self} and so on. Similar methods have been applied to medical images \cite{jamaludin2017self,zhou2019models,bai2019self,chen2019self}. However, most of these works still require a second-stage fine-tuning after initializing with weights learned from self-supervision. In addition, features learned from handcrafted tasks may not be sufficiently generalizable to semantic segmentation, as two tasks might not be strongly related \cite{zamir2018taskonomy}. In contrast, in our work, segmenting superpixel-based pseudolabels is directly related to segmenting real objects. This is because superpixels are compact building blocks for semantic masks for real objects. Recent works \cite{dou2019domain,yu2020foal,wu2019deep} on medical imaging rely on second-order optimization \cite{finn2017model}. These works differ from our work in key method and task. 

Our proposed SSL technique shares a similar spirit as \cite{ji2019invariant} (or arguably, as some recent works on contrastive learning \cite{wu2018unsupervised,bachman2019learning,he2020momentum,chen2020improved}) in methodology. Both methods encourage invariance in image representation by intentionally creating variants. While \cite{ji2019invariant} focuses on visual information clustering, we focus on the practical but challenging few-shot medical image segmentation problem. 

\subsection{Superpixel segmentation}
\label{subsec: superpix}
Superpixels are small, compact image segments which are usually piece-wise smooth \cite{mumford1989optimal,ren2003learning}. Superpixels are generated by clustering local pixels using statistical models with respect to low-level image features. These models include Gaussian mixture \cite{achanta2010slic} and graph cut \cite{liu2011entropy}. 
In this work, we employed off-the-shelf, efficient and unsupervised graph-cut-based algorithm by \cite{felzenszwalb2004efficient}. Compared with the popular SLIC method \cite{achanta2010slic}, superpixels generated by \cite{felzenszwalb2004efficient} are more diverse in shape. Training with these superpixels intuitively improves generalizability of the network to unseen classes in various shapes. %We refer readers to \cite{stutz2018superpixels} for a detailed review of superpixels. 

\section{Method}
\label{sec: method}
We first introduce problem formulation for few-shot semantic segentation (FSS). Then, the ALPNet architecture is introduced with a focus on adaptive local prototype pooling and the corresponding inference process. We highlight our superpixel-based self-supervised learning (SSL) with details in pseudolabel generation process and episode formation in Section \ref{subsec: s3}. Finally, we introduce the overall end-to-end training objective under the proposed SSL technique. Of note, after the proposed self-supervised learning, ALPNet can be directly applied to unseen classes with its weights fixed, and with reference to a few human-labeled support slices. There is no fine-tuning required in this testing phase.

\subsection{Problem Formulation}
\label{subsec: problem}
The aim of few-shot segmentation is to obtain a model that can segment an unseen semantic class, by just learning from a few labeled images of this unseen class during inference without retraining the model.
In few-shot segmentation, a training set $\mcl{D}_{tr}$ containing images with training semantic classes $\mcl{C}_{tr}$ (e.g., $\mcl{C}_{tr} = \{ \textit{liver}, \textit{spleen}, \textit{spine} \}$), and a testing set $\mcl{D}_{te}$ of images containing testing unseen classes $\mcl{C}_{te}$ (e.g., $\mcl{C}_{te} = \{ \textit{heart}, \textit{kidney} \}$), are given, where $\mcl{C}_{tr} \cap \mcl{C}_{te} = \text{\O}$. The task is to train a segmentation model on $\mcl{D}_{tr}$ (e.g. labeled images of livers, spleens and spines) that can segment semantic classes $\mcl{C}_{te}$ in images in $\mcl{D}_{te}$, given a few annotated examples of $\mcl{C}_{te}$ (e.g. to segment \textit{kidney} with reference to a few labeled images of kidney), without re-training. 
$\mcl{D}_{tr} = \{ (\mbx, \mby(\cjh)) \}$ is composed of images $\mbx \in \mcl{X}$ and corresponding binary masks $\mby(\cjh)\text{'s} \in \mcl{Y}$ of classes $\cjh \in \mcl{C}_{tr}$, where $\hat{j} = 1,2,3,...,N$ is the class index. $\mcl{D}_{te}$ is defined in the same way but for testing images and masks with $\mcl{C}_{te}$. 
\begin{figure}[!t]
\centering
\includegraphics[width=12cm]{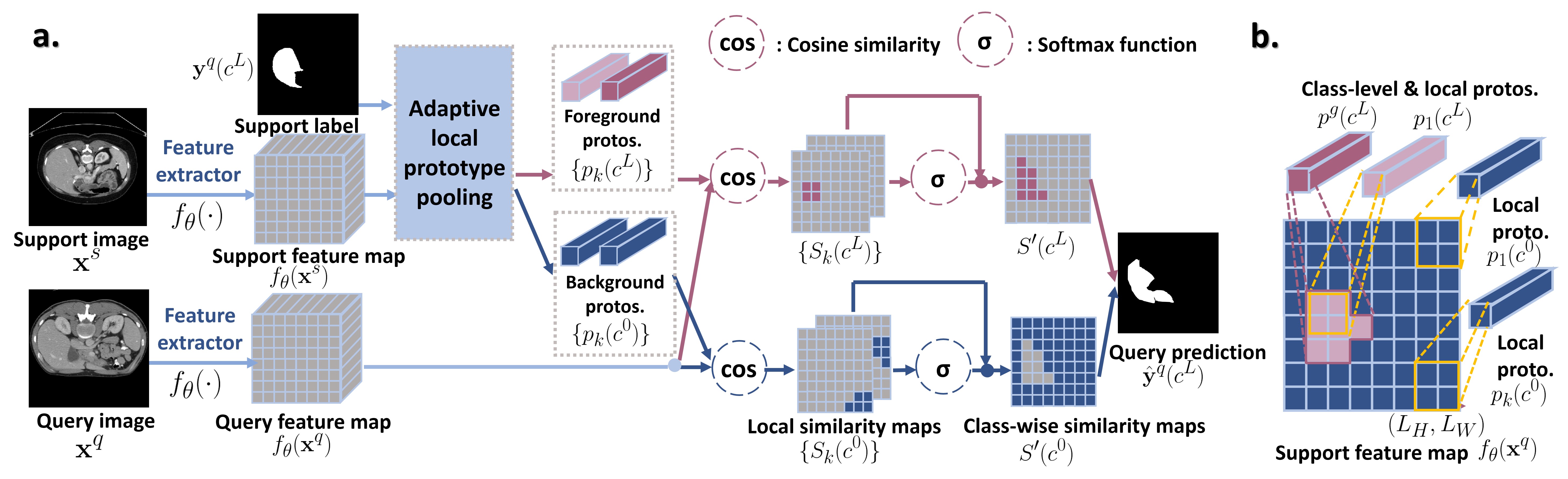}
\caption{(a). Workflow of the proposed network: The feature extractor $f_{\theta}(\cdot)$ takes the support image and query image as input to generate feature maps $f_{\theta}(\mbx^s)$ for support and $f_{\theta}(\mbx^q)$ for query. The proposed adaptive local prototype pooling module then takes support feature map and support label as input to obtain an ensemble of representation prototypes $p_{k}(\cj)$'s. These prototypes are used as references for comparing with query feature map $f_{\theta}(\mbx^q)$. Similarity maps generated by these comparisons are fused together to form the final segmentation. This figure illustrates a 1-way segmentation setting, where $c^L$ is the foreground class, $c^0$ is the background. (b). Illustration of the adaptive local prototype pooling module: Local prototypes are calculated by spatially averaging support feature maps within pooling windows (orange boxes); class-level prototypes are averaged under the entire support label (purple region).}
\label{fig:network}
\end{figure}
In each inference pass, a \textit{support} set $\mcl{S}$ and a \textit{query} set $\mcl{Q}$ are given. The support $\mcl{S} = \{ (\mbx^s_l, \mby^s_l(\cjh) ) \}$ contains images $\mbx^s_l$'s and masks $\mby^s_l(\cjh)$'s, and it serves as examples for segmenting $\cjh$'s; the query set $\mcl{Q} = \{ \mbx^q \}$ contains images $\mbx^q$'s to be segmented. Here, the superscripts denote an image or mask is from support ($s$) or query ($q$). And $l=1,2,3,...,K$ is the index for each image-mask pair of class $\cjh$. One support-query pair $( \mcl{S}, \mcl{Q} )$ comprises an $episode$. Every episode defines an $N$-way $K$-shot segmentation sub-problem if there are $N$ classes (also called $N$ tasks) to be segmented and $K$ labeled images in $\mcl{S}$ for each class. Note that the background class is denoted as $c^0$ and it does not count toward $\mcl{C}_{tr}$ or $\mcl{C}_{te}$. 
\subsection{Network Architecture}
\label{subsec: pn}
 \noindent \textbf{Overview}: Our network is composed of: (a) a generic \textit{feature extractor} network $f_{\theta}(\cdot): \mcl{X} \xrightarrow{} \mcl{E}$ parameterized by $\theta$, where $\mcl{E}$ is the representation space (i.e. feature space) on which segmentation operates; (b) the proposed \textit{adaptive local prototype pooling module} (ALP) $g(\cdot,\cdot): \mcl{E}\times \mcl{Y} \xrightarrow{} \mcl{E}$ for extracting representation \textit{prototypes} from support features and labels; (c) and a \textit{similarity based classifier} $sim(\cdot, \cdot): \mcl{E}\times \mcl{E} \xrightarrow{} \mcl{Y}$ for segmention by comparing prototypes and query features. 

As shown in Fig. \ref{fig:network}, in inference, the feature extractor network $f_{\theta}(\cdot)$ provides ALP with feature maps by mapping both $\mbx^s_l$'s and $\mbx^q$'s to feature space $\mcl{E}$, producing feature maps $\{(f_{\theta}(\mbx^q), f_{\theta}(\mbx^s_l))\} \in \mcl{E}$. ALP takes each $(f_{\theta}(\mbx^s_l), \mby^s_l(\cjh))$ pair as input to compute both \textit{local prototypes} and \textit{class-level prototypes} of semantic class $\cjh$ and background $c^0$. These prototypes will later be used as references of each class for segmenting query images. Prototypes of all $\cj$'s forms a prototype ensemble $\mcl{P} = \{ p_k(\cj) \}, j=0,1,2,...,N$ where $k$ is prototype index and $k \geq 1$ for each $\cj$. %In \cite{wang2019panet} this is done by \textit{masked average pooling} (MAP): averaging $f_{\theta}(\cdot)$ under $\mby^s(c)$. 
This prototype ensemble is used by the classifier $sim(\cdot, \cdot)$ to predict the segmentation for the query image, saying $\hat{\mby}^q = sim(\mcl{P}, f_{\theta}(\mbx^q))$. This is achieved by first measuring similarities between each $p_k(\cj)$'s and query feature map $f_{\theta}(\mbx^q)$, and then fusing these similarities together. \\
\noindent \textbf{Adaptive local prototype pooling}: In contrast to previous works \cite{wang2019panet,siam2019amp,liu2020prototype}, where intra-class local information is unreasonably spatially averaged out underneath the semantic mask, we propose to preserve local information in prototypes by introducing adaptive local prototype pooling module (ALP). In ALP, each \textit{local prototype} is only computed within a \textit{local pooling window} overlaid on the support and only represents one part of object-of-interest.

Specifically, we perform average pooling with a pooling window size $(L_H, L_W)$ on each $f_{\theta}(\mbx^s_l) \in \mathbb{R}^{D \times H \times W}$ where $(H,W)$ is the spatial size and $D$ is the channel depth. Of note, $(L_H, L_W)$ determines the spatial extent under which each local prototype is calculated in the representation space $\mcl{E}$. The obtained local prototype $p_{l,mn}(c)$ with undecided class $c$ at spatial location $(m,n)$ of the average-pooled feature map is given by
\begin{small}
\begin{align}
\label{equ: local_proto_calc}
p_{l,mn}(c) = \text{avgpool}(f_{\theta}(\mbx^s_l))(m,n) = \frac{1}{L_H L_W} \sum_{h} \sum_{w} f_{\theta}(\mbx^s_l)(h,w), \\ \nonumber
\text{where} \ mL_H \leq h < (m+1) L_H, \ nL_W \leq w < (n+1)L_W.
\end{align}
\end{small}
To decide the class $c$ of each $p_{l,mn}(c)$, we average-pool the binary mask $\mby^s_l(\cjh)$ of the foreground class $\cjh$ to the same size $(\frac{H}{L_H}, \frac{W}{L_W})$. Let $y^a_{l,mn}$ be the value of $\mby^s_l(\cjh)$ after average pooling at location $(m,n)$, $c$ is assigned as:
\begin{small}
\begin{align}
\label{equ: class_of_proto}
c =\begin{cases} c^0 & y^a_{l,mn} < T \\
                     \cjh &  y^a_{l,mn} \geq T
       \end{cases} \ \ \text{where} \ \ 
y^a_{l,mn} = \text{avgpool}(\mby^s_l(\cjh))(m,n).
\end{align}
\end{small}
$T$ is the lower-bound threshold for foreground which is empirically set to 0.95. 

To ensure at least one prototype is generated for objects smaller than the pooling window $(L_H, L_W)$, we also compute a \textit{class-level prototype} $p^{g}_l(\cjh)$ using masked average pooling \cite{zhang2018sg,wang2019panet}:%, which is not confined by window size of average pooling.
\begin{small}
\begin{align}
\label{equ: global_proto}
p^{g}_l(\cjh) = \frac{ \underset{h,w}{\sum}  \mby^s_l(\cjh)(h,w) f_{\theta}(\mbx^s_l)(h,w) }{\underset{h,w}{\sum} \mby_l^s(\cjh)(h,w)     }.
\end{align}
\end{small}
In the end, $p_{l,mn}(\cj)$'s and $p_l^g(\cjh)$'s are re-indexed with subscript $k$'s for convenience, and hence comprise the representation prototype ensemble $\mcl{P} = \{p_k(\cj)\}$. This ensemble therefore preserves more intra-class local distinctions by explicitly representing different local regions into separate prototypes.

\noindent \textbf{Similarity-based segmentation}: The similarity-based classifier $sim(\cdot, \cdot)$ is designed to make dense prediction on query by exploiting local image information in $\mcl{P}$. This is achieved by firstly matching each prototype to a corresponding local region in query, and then fusing the local similarities together. 

As a loose interpretation, to segment a large \textit{liver} in query, in the first stage, a local prototype $p_k(c^L)$ with class $c^L=\textit{liver}$, whose pooling window falls over the \textit{right lobe} of the \textit{liver} particularly finds a similar region which looks like a \textit{right lobe} in query (instead of matching the entire \textit{liver}). Then, to get an entire liver, results from \textit{right lobe}, and \textit{left lobe} are fused together to form a \textit{liver}.

Specifically, $sim(\cdot, \cdot)$ first takes query feature map $f_{\theta}(\mbx^q)$ and prototype ensemble $\mcl{P} = \{p_k(\cj)\}$ as input to compute \textit{local similarity maps} $S_k(\cj)$'s between $f_{\theta}(\mbx^q)$ and all $p_k(\cj)$'s respectivelyy. Each entry $S_k(\cj)(h, w)$ at spatial location $(h,w)$ corresponding to $f_{\theta}(\mbx^q)$ is given by 
\begin{small}
\begin{align}
\label{equ: local_proto_sim}
S_k(\cj)(h,w) = \alpha p_k(\cj) \odot f_{\theta}(\mbx^q)(h,w),
\end{align}
\end{small}
where $\odot$ denotes cosine similarly, which is bounded, same as in \cite{wang2019panet}: $a \odot b = \frac{ \langle a,b \rangle }{ \Vert a  \Vert_2 \Vert b \Vert_2 }, \ a,b \in \mathbb{R}^{D \times 1 \times 1}$, $\alpha$ is a multiplier, which helps gradients to back-propagate in training \cite{oreshkin2018tadam}.  In our experiments, $\alpha$ is set to 20, same as in \cite{wang2019panet}.

Then, to obtain similarity maps (unnormalized) with respect to each class $\cj$ as a whole, local similarity maps $S_k(\cj)$'s are fused for each class separately into \textit{class-wise similarities} $S'(\cj)$, this is done through a softmax function:
\begin{small}
\begin{align}
\label{equ: global_prob}
S'(\cj)(h,w) = &\underset{k}{\sum} S_k(\cj)(h,w) \ \underset{k}{ \text{softmax}}[S_k(\cj)(h,w)].
\end{align}
\end{small}
$\underset{k}{ \text{softmax}}[S_k(\cj)(h,w)]$ refers to the operation of first stacking all $S_k(\cj)(h,w)$'s along channel dimension and then computing softmax function along channels.

To obtain the final dense prediction, in the end, class-wise similarities are normalized into probabilities:
\begin{small}
\begin{align}
\label{equ: pred_prob}
\hat{\mby}^q(h,w) = \underset{j}{\text{softmax}}[S'(\cj)(h,w) ].
\end{align}
\end{small}
\subsection{Superpixel-based Self-supervised Learning}
\label{subsec: s3}
\begin{figure}[!htbp]
\centering
\includegraphics[width=11cm]{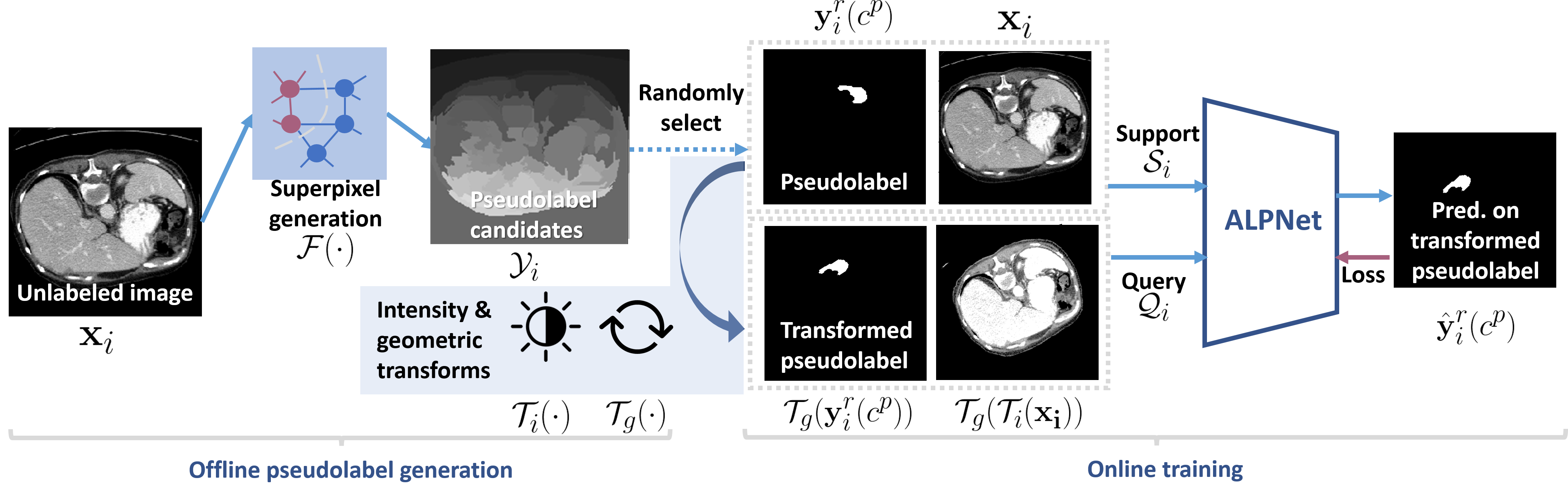}
\caption{Workflow of the proposed superpixel-based self-supervised learning technique.}
\label{fig:selfsup}
\end{figure}
To obtain accurate and robust results, two properties are highly desirable for similarity-based classifiers. 
For each class, the representations should be \textit{clustered} in order to be discriminative under a similarity metric; meanwhile, these representations should be \textit{invariant} across images (in our case any combinations of support and query) to ensure robustness in prediction \cite{ji2019invariant}. 

These two properties are encouraged by the proposed superpixel-based self-supervised learning (SSL). 
As annotations for real semantic classes are unavailable, SSL exploits pseudolabels to enforce \textit{clustering} at a superpixel-level. This is naturally achieved by back-propagating segmentation loss via cosine-similarity-based classifier. Here, the superpixel-level clustering property can be  transferred to real semantic classes, since one semantic mask is usually composed of several superpixels \cite{stutz2018superpixels,ren2003learning}. Additionally, to encourage representations to be invariant against shape and intensity differences between images, we perform geometric and intensity transforms between support and query. This is because shape and intensity are the largest sources of variations in medical images \cite{heimann2009statistical}. 

The proposed SSL framework consists of two phases: offline pseudolabel generation and online training. The entire workflow can be seen in Fig. \ref{fig:selfsup}.\\
\noindent \textbf{Unspervised pseudolabel generation}:
To obtain candidates for pseudolabels, a collection of superpixels $\mathcal{Y}_i = \mcl{F}(\mbx_i)$ are generated for every image $\mbx_i$. This is efficiently done with the unsupervised algorithm \cite{felzenszwalb2004efficient} denoted by $\mcl{F} (\cdot)$. 

\noindent\textbf{Online episode composition}: For each episode $i$, an image $\mbx_i$ and a randomly chosen superpixel $\mby_i^r(c^p) \in \mcl{Y}_i$ form the support $\mcl{S}_i = \{(\mbx_i, \mby^r_i(c^p))\}$. Here $\mby_i^r(c^p)$ is a binary mask with index $r=1,2,3,... ,|\mcl{Y}_i|\text{ and } c^p$ denotes the \textit{pseudolabel} class (corresponding background mask $\mby_i^r(c^0)$ is given by $1 - \mby_i^r(c^p)$). Meanwhile, the query set $\mcl{Q}_i = \{( \mcl{T}_g(\mcl{T}_i(\mbx_i))) \}, \mcl{T}_g (\mby_i^r(c^p) ))$ is constructed by applying random geometric and intensity transforms: $\mcl{T}_g(\cdot)$ and $\mcl{T}_i(\cdot)$ to the support. By this mean, each $(\mcl{S}_i, \mcl{Q}_i)$ forms a 1-way 1-shot segmentation problem. In practice, $\mcl{T}_g(\cdot)$ includes affine and elastic transforms, $\mcl{T}_i(\cdot)$ is gamma transform.

\noindent \textbf{End-to-end training}: The network is trained end-to-end, where each iteration $i$ takes an episode $(\mcl{S}_i, \mcl{Q}_i)$ as input. Cross entropy loss is employed where the segmentation loss $\mcl{L}_{seg}^i$ for each iteration is written as:
%  with episodes $(\mcl{S}_i, \mcl{Q}_i)$'s.

\begin{small}
\begin{align}
\label{equ: loss_seg}
\mcl{L}^i_{seg}(\theta ; \mcl{S}_i, \mcl{Q}_i) = - \frac{1}{HW} \sum_{h}^H \sum_{w}^W \sum_{j \in \{0,p\} }
\mathcal{T}_g(\mby^r_i(c^j))(h,w) \log( \hat{\mby}^r_i(c^j)(h,w) ),
\end{align}
\end{small}

where $\hat{\mby}^r_i(c^p)$ is the prediction of query pseudolabel $\mcl{T}_g (\mby_i^r(c^p) )$ and is obtained as described in Section \ref{subsec: pn}. In practice, weightings of 0.05 and 1.0 are given to $c^0$ and $c^p$ separately for mitigating class imbalance.
We also inherited the \textit{prototypical alignment regularization} in \cite{wang2019panet}: taking prediction as support, i.e. $\mcl{S}' =(\mcl{T}_g( \mcl{T}_i(\mbx_i )),\hat{\mby}^r_i(c^p))$, it should correctly segment the original support image $\mbx_i$. This is presented as
\begin{small}
\begin{align}
\label{equ: loss_reg}
\mcl{L}^i_{reg}(\theta ; \mcl{S}'_i, \mcl{S}_i) = - \frac{1}{HW} \sum_{h}^H \sum_{w}^W \sum_{j \in \{0,p\} } \mby^r_i(c^j)(h,w) \log( \bar{\mby}^r_i(c^j)(h,w) ),
\end{align}
\end{small}
where $\bar{\mby}^r_i(c^p)$ is the prediction of $\mby^r_i(c^p)$ taking $\mbx_i$ as query. % by taking $\mcl{S}' =(\mcl{T}_g(\mbx_i),\bar{\mby}^r_i(c^p))$ as support. 

Overall, the loss function for each epioside is:
\begin{small}
\begin{align}
\label{equ: overall}
\mcl{L}^i(\theta; \mcl{S}_i, \mcl{Q}_i) =\mcl{L}^i_{seg} + \lambda \mcl{L}^i_{reg},
\end{align}
\end{small}
where $\lambda$ control strength of regularization as in \cite{wang2019panet}.  

After self-supervised learning, the network can be directly used for inference on unseen classes.
\section{Experiments}
\noindent \textbf{Datasets}: 
To demonstrate the general applicability of our proposed method under different imaging modalities, segmentation classes and health conditions of the subject, we performed evaluations under three scenarios: abdominal organs segmentation for CT and MRI (Abd-CT and Abd-MRI) and cardiac segmentation for MRI (Card-MRI). 
All three datasets contain rich information outside their regions-of-interests, which benefits SSL by providing sources of superpixels.
Specifically,
\begin{itemize}
    \item \textbf{Abd-CT} is from MICCAI 2015 Multi-Atlas Abdomen Labeling challenge \cite{landman2015miccai}. It contains 30 3D abdominal CT scans. Of note, this is a clinical dataset containing patients with various pathologies and variations in intensity distributions between scans.
    \item \textbf{Abd-MRI} is from ISBI 2019 Combined Healthy Abdominal Organ Segmentation Challenge (Task 5)  \cite{kavur2020chaos}. It contains 20 3D T2-SPIR MRI scans. 
    \item \textbf{Card-MRI} is from MICCAI 2019 Multi-sequence Cardiac MRI Segmentation Challenge (bSSFP fold) \cite{zhuang2018multivariate}, with 35 clinical 3D cardiac MRI scans.% This is a also a clinical dataset.
\end{itemize}

To unify experiment settings, all images are re-formated as 2D axial (Abd-CT and Abd-MRI) or 2D short-axis (Card-MRI) slices, and resized to $256\times256$ pixels. Prepossessings are applied following common practices. Each 2D slice is repeated for three times in channel dimension to fit into the network.

To comparatively evaluate the results on classes with various shapes, locations and textures between partically-pathologic, imhomogeneous Abd-CT and all-healthy, homogeneous Abd-MRI, we construct a shared label set containing left kidney, right kidney, spleen and liver; For Card-MRI, the label set contains left-ventricle blood pool (LV-BP), left-ventricle myocardium (LV-MYO) and right-ventricle (RV). In all experiments, we perform five-fold cross-validation.

\noindent \textbf{Evaluation}:
To measure the overlapping between prediction and ground truth, we employ Dice score (0-100, 0: mismatch; 100: perfectly match), which is commonly used in medical image segmentation researches. 
To evaluate 2D segmentation on 3D volumetric images, we follow the evaluation protocol established by \cite{roy2020squeeze}. In a 3D image, for each class $\cjh$, images between the top slice and the bottom slice containing $\cjh$ are divided into $C$ equally-spaced chunks. The middle slice in each chunk from the support scan is used as reference for segmenting all the slices in corresponding chunk in query. In our experiments $C$ is set to be 3. Of note, the support and query scans are from different patients.

To evaluate generalization ability to unseen testing classes, beyond the standard few-shot segmentation experiment setting for medical images established by \cite{roy2020squeeze} (\textbf{setting 1}), where testing class might appear as background in training data, we introduce a \textbf{setting 2}. In setting 2, we force testing classes (even unlabeled) to be completely unseen by removing any image that contains a testing class, from the training dataset. 

Labels are therefore partitioned differently according to the settings and types of supervision. In setting 1, when training with SSL, no label partitioning is required for training. 
When training with annotated images, each time we take one class for testing and the rest for training. To observe if the learned representations encode spatial concepts like left and right, we deliberately group $\langle$left/ right kidney$\rangle$ to appear together in training or testing. In setting 2, as $\langle$spleen, liver$\rangle$, or $\langle$left/ right kidney$\rangle$ usually appear together in a 2D slice respectively, we group them into \textit{upper} abdomen and \textit{lower} abdomen groups separately. In each experiment all slices containing the testing group will be removed from training data. For Card-MRI, only setting 1 is examined as all the labels usually appear together in 2D slices, making label exclusion impossible. 

To simulate the scarcity of labeled data in clinical practice, all our experiments in this section are performed under 1-way 1-shot setting.

\noindent \textbf{Implementation Details}:
The network is implemented with PyTorch based on official PANet implementation\footnote{https://github.com/kaixin96/PANet} \cite{wang2019panet}. To obtain high spatial resolutions in feature maps, $f_{\theta}(\cdot)$ is configured as an off-the-shelf fully-convolutional ResNet101, which is pre-trained on part of MS-COCO for higher segmentation performance \cite{wang2019panet,shin2016deep} (same for vanilla PANet in our experiments). 
It takes a $3 \times 256 \times 256$ image  as input and produces a $256\times32\times32$ feature map. Local pooling window $(L_H, L_W)$ for prototypes  is set to $4\times4$ for training and $2\times2$ for inference on feature maps. The loss in Equ. \ref{equ: overall} is minimized for 100k iterations using stochastic gradient descent with a batch size of 1. The learning rate is 0.001 with a stepping decay rate of 0.98 per 1000 iterations. The self-supervised training takes $\sim$3h on a single Nvidia RTX 2080Ti GPU, consuming 2.8GBs of memory. 
\subsection{Quantitative and qualitative Results}
\begin{figure}[!t]
\centering
\includegraphics[width=11.5cm]{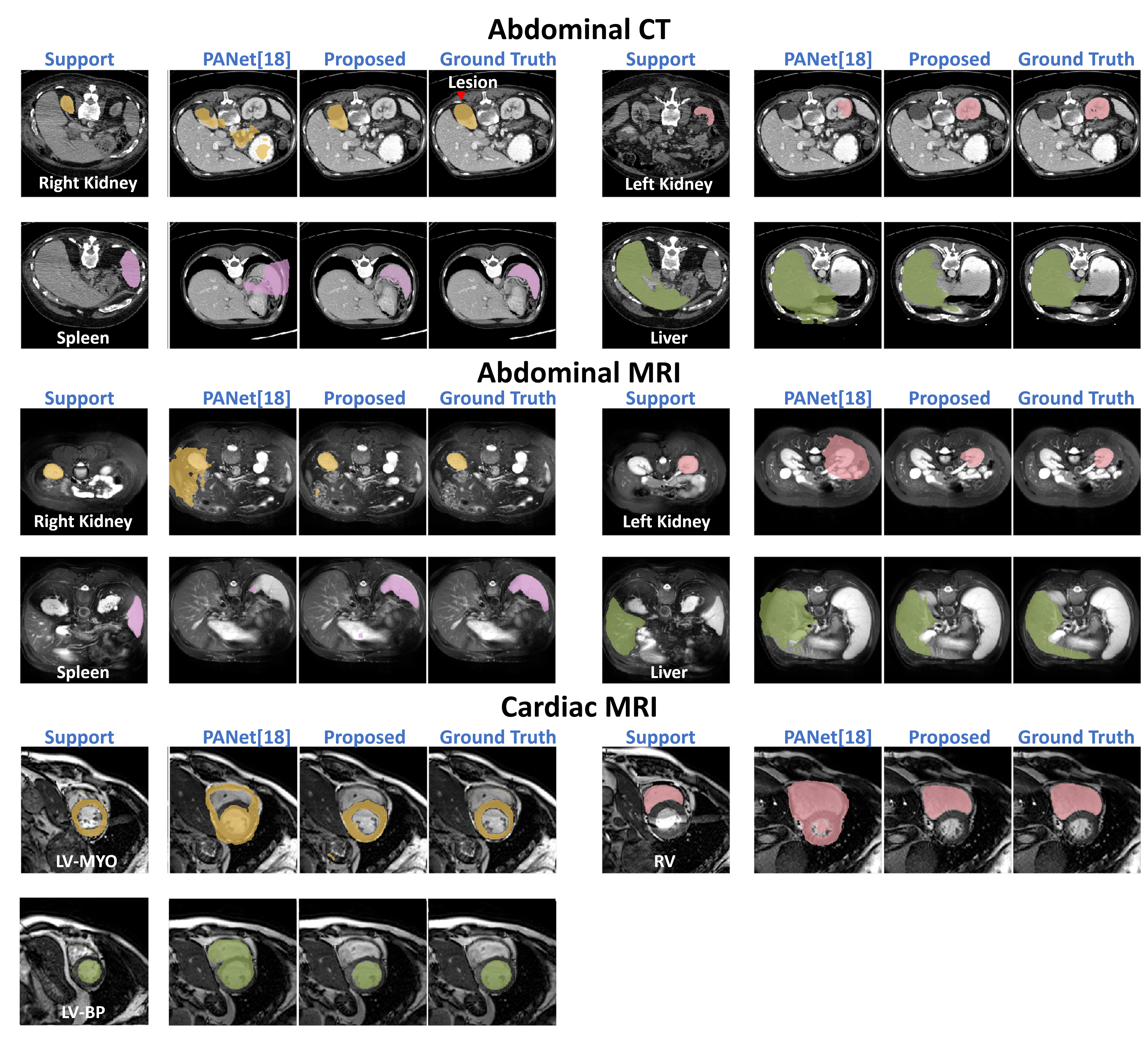}
\caption{Qualitative results of our method on all three combinations of imaging modalities and segmentation tasks. The proposed method achieves desirable segmentation results which are close to ground truth. To highlight the strong generalization ability, examples of results from the proposed method on Abd-CT and Abd-MRI are from setting 2, where images containing testing classes are strictly excluded in training set even though they are unlabeled. See supplemental materials for more examples.}
\label{fig: segmentation}
\end{figure}

\noindent \textbf{Comparison with state-of-the-art methods}:
\begin{table}[ht]
\centering
\caption{Experiment results (in Dice score) on abdominal images under setting 2.}
\resizebox{0.85\linewidth}{!}{%
\begin{tabular}{lcYYYYYYYYYY}
\toprule
\multirow{3}{*}{Method} & \multirow{3}{*}{Manual Anno.?} & \multicolumn{5}{c}{Abdominal-CT} & \multicolumn{5}{c}{Abdominal-MRI} \\
& & \multicolumn{2}{c}{Lower} & \multicolumn{2}{c}{Upper} & \multirow{2}{*}{Mean} & \multicolumn{2}{c}{Lower} & \multicolumn{2}{c}{Upper} & \multirow{2}{*}{Mean}  \\
& & LK & RK & Spleen & Liver & & LK & RK & Spleen & Liver & \\ 
\cmidrule(lr){1-1}\cmidrule(lr){2-2}\cmidrule(lr){3-4}\cmidrule(lr){5-6}\cmidrule(lr){7-7}\cmidrule(lr){8-9}\cmidrule(lr){10-11}\cmidrule(lr){12-12}
SE-Net \cite{roy2020squeeze} & \checkmark & 32.83  & 14.34  & 0.23  & 0.27  & 11.91 & 62.11 & 61.32 &51.80   & 27.43  & 50.66  \\
Vanilla PANet \cite{wang2019panet} & \checkmark & 32.34  & 17.37  & 29.59 & 38.42  & 29.43 & 53.45  & 38.64  &50.90  & 42.26  & 46.33  \\
\midrule
ALPNet-init & - & 13.90 & 11.61  & 16.39 & 41.71 & 20.90 & 19.28 & 14.93 & 23.76  & 37.73 & 23.93  \\
ALPNet & \checkmark & 34.96  & 30.40  & 27.73  & 47.37  & 35.11  & 53.21  & 58.99  &52.18   & 37.32  & 50.43  \\
SSL-PANet & $\times$ & 37.58  & 34.69 & 43.73 & 61.71 & 44.42 & 47.71 & 47.95  & 58.73  & 64.99 & 54.85 \\
SSL-ALPNet & $\times$ & \textbf{63.34}  & \textbf{54.82}  & \textbf{60.25} & \textbf{73.65}  & \textbf{63.02}  & \textbf{73.63}  & \textbf{78.39}  &\textbf{67.02}  & \textbf{73.05} & \textbf{73.02} \\
\bottomrule
\end{tabular}
}
\label{tbl: strictly_inductive_few_shot}
\end{table}
\begin{table}[ht]
\centering
\caption{Experiment results (in Dice score) on abdominal images under setting 1.}
\resizebox{0.85\linewidth}{!}{%
\begin{tabular}{lcYYYYYYYYYY} 
\toprule
\multirow{3}{*}{Method} & \multirow{3}{*}{Manual Anno.?} & \multicolumn{5}{c}{Abdominal-CT} & \multicolumn{5}{c}{Abdominal-MRI} \\
                     &  & \multicolumn{2}{c}{Kidneys} & \multirow{2}{*}{Spleen} & \multirow{2}{*}{Liver} & \multirow{2}{*}{Mean} & \multicolumn{2}{c}{Kidneys}  & \multirow{2}{*}{Spleen} & \multirow{2}{*}{Liver} & \multirow{2}{*}{Mean} \\
                     &  & LK & RK &  & & & LK & RK & & & \\ 
\cmidrule(lr){1-1}\cmidrule(lr){2-2}\cmidrule(lr){3-4}\cmidrule(lr){5-5}\cmidrule(lr){6-6}\cmidrule(lr){7-7}\cmidrule(lr){8-9}\cmidrule(lr){10-10}\cmidrule(lr){11-11}\cmidrule(lr){12-12}
SE-Net \cite{roy2020squeeze} & \checkmark & 24.42  & 12.51   & 43.66  & 35.42  & 29.00  & 45.78  & 47.96 & 47.30  & 29.02  & 42.51   \\
Vanilla PANet \cite{wang2019panet} & \checkmark & 20.67  & 21.19   & 36.04  & 49.55  & 31.86  & 30.99  & 32.19  & 40.58  & 50.40  & 38.53   \\
\midrule
ALPNet & \checkmark & 29.12  & 31.32  & 41.00 & 65.07  & 41.63  & 44.73  & 48.42 & 49.61 & 62.35  & 51.28 \\
SSL-PANet & $\times$ & 56.52 & 50.42  & 55.72 & 60.86  & 57.88  & 58.83  & 60.81  & 61.32  & 71.73  & 63.17  \\
SSL-ALPNet & $\times$ & \textbf{72.36}  & \textbf{71.81}   & \textbf{70.96}  & \textbf{78.29}  & \textbf{73.35}  & \textbf{81.92} & \textbf{85.18} & \textbf{72.18}  &  \textbf{76.10} & \textbf{78.84}  \\
\midrule
Zhou et al. \cite{zhou2019prior} & Ful. Sup. & 95.3 & 92.0 & 96.8 & 97.4 & 95.4 & \multicolumn{5}{c}{-} \\
Isenseen et al. \cite{isensee2018nnu} & Ful. Sup. & \multicolumn{5}{c}{-} &  & \multicolumn{3}{c}{-} & 94.6  \\
\bottomrule
\end{tabular}
}
\label{tbl: relaxed_few_shot}
\end{table}
\begin{table}[!h]
\centering
\caption{Experiment results (in Dice score) on cardiac images under setting 1.}
\resizebox{0.5\linewidth}{!}{%
\begin{tabular}{lccccc} 
\toprule
Method & Manual Anno.? & LV-BP & LV-MYO & RV & Mean  \\ 
\cmidrule(lr){1-1}\cmidrule(lr){2-2}\cmidrule(lr){3-3}\cmidrule(lr){4-4}\cmidrule(lr){5-5}\cmidrule(lr){6-6}
SE-Net \cite{roy2020squeeze} & \checkmark & 58.04 & 25.18 & 12.86 & 32.03 \\
Vanilla PANet \cite{wang2019panet} & \checkmark & 53.64 & 35.72  & 39.52 & 42.96 \\
\midrule
%cmidrule(lr){1-1}
ALPNet & \checkmark & 73.08 & 49.53 & 58.50 & 60.34  \\
SSL-PANet & $\times$ & 70.43 & 46.79 & 69.52 & 62.25  \\
SSL-ALPNet  & $\times$ & \textbf{83.99}  & \textbf{66.74}  & \textbf{79.96} & \textbf{76.90}  \\
\bottomrule
\end{tabular}
}
\label{tbl: relaxed_few_show_cardiac}
\end{table}
Table \ref{tbl: strictly_inductive_few_shot} - \ref{tbl: relaxed_few_show_cardiac} show the comparisons of our method with vanilla PANet, one of state-of-the-art methods on natural images and SE-Net\footnote{https://github.com/abhi4ssj/few-shot-segmentation} \cite{roy2020squeeze}, the lastest FSS method for medical images. Without using any manual annotation, our proposed SSL-ALPNet consistently outperforms them by an average Dice score of $>$25. As shown in Fig. \ref{fig: segmentation}, the proposed framework yields satisfying results on organs with various shapes, sizes and intensities. Of note, for all evaluated methods, results on Abd-MRI are in general higher than those on Abd-CT. This not suprising as Abd-MRI is more homogeneous, and most of organs in Abd-MRI have distinct contrast to surrounding tissues, which helps to reduce ambiguity at boundaries.
 
Importantly, Table \ref{tbl: strictly_inductive_few_shot} demonstrates the strong generalization ability of our method to unseen classes. 
This implies that the proposed superpixel-based self-supervised learning has successfully trained the network to learn more diverse and generalizable image representations from unlabeled images.

The upperbounds obtained by fully-supervised learning on all labeled images are shown in Table \ref{tbl: relaxed_few_shot} for reference.

\noindent \textbf{Performance boosts by ALP and SSL}:
The separate performance gains obtained by introducing adaptive local prototype pooling or self-supervised learning can be observed in rows \textit{ALPNet} and \textit{SSL-PANet} in Table \ref{tbl: strictly_inductive_few_shot} - \ref{tbl: relaxed_few_show_cardiac}.
These results suggests both SSL and ALP contribute greatly. The performance gains of SSL highlight the benefit of a well-designed training strategy that encourages learning generalizable features, which is usually overlooked in recent few-shot segmentation methods. More importantly, the synergy between them (\textit{SSL-ALPNet}) leads to significant performance gains by learning richer image representations and by constructing more effective inductive bias. To be assured that MS-COCO initialization alone cannot do FSS, we also include the results when the ALPNet is directly tested after initialization, shown in \textit{ALPNet-init} in Table \ref{tbl: strictly_inductive_few_shot}.

\noindent \textbf{Robustness under patient pathology}: As shown in Fig \ref{fig: segmentation}, despite the large dark lesion on right-kidney in Abd-CT, the proposed method stably produces satisfying results.

\subsection{Ablation studies}
Ablation studies are performed on Abd-CT under setting 2. This scenario is challenging but close to clinical scenario in practice.

\noindent \textbf{Importance of transforms between the support and query}: To demonstrate the importance of geometric and intensity transformations in our method, we performed ablation studies as shown in Table. \ref{tbl: transforms_support_query}. Unsurprisingly, the highest and lowest overall results are obtained by applying both or no transforms, proving the effectiveness of introducing random transforms. 
\begin{table}[h]
\begin{minipage}{.49\textwidth}

\caption{Ablation study on types transformations.}
\resizebox{0.85\linewidth}{!}{%
\begin{tabular}{ccccccc} 
\toprule
Int. & Geo. & LK & RK & Spleen & Liver & Mean\\ 
\cmidrule(lr){1-1}\cmidrule(lr){2-2}\cmidrule(lr){3-4}\cmidrule(lr){5-6}\cmidrule(lr){7-7}
$\times$ & $\times$ & 45.49 & 48.40 & 53.05 & 73.60 & 55.13  \\
\checkmark & $\times$ & 55.56  & 49.12 & 59.20 & 73.39 &  59.31 \\
$\times$ & \checkmark & 59.32 & 51.45 & 57.74 & \textbf{78.93} & 61.86 \\
\checkmark & $\checkmark$ & \textbf{63.34} & \textbf{54.82} & \textbf{60.25} & 73.65  & \textbf{63.02} \\
\bottomrule
\end{tabular}
}
\label{tbl: transforms_support_query}
\end{minipage}
\hfill
\begin{minipage}{.49\textwidth}
\caption{Ablation study on minimum pseudolabel sizes.}
\resizebox{1.0\linewidth}{!}{%
\begin{tabular}{lccccc} 
\toprule
Min. size (px) & LK & RK & Spleen & Liver & Mean  \\ 
\cmidrule(lr){1-1}\cmidrule(lr){2-3}\cmidrule(lr){4-5}\cmidrule(lr){6-6}
100 & 52.92 & 47.45 & 53.16 & 68.40 & 55.49 \\
400 & \textbf{63.34} & \textbf{54.82} & \textbf{60.25} & 73.65 & \textbf{63.02}  \\
1600 & 51.74 & 44.83  & 56.99  & \textbf{74.73} & 57.08   \\
\midrule
Avg. Size in 2D (px) & 798 & 799 & 1602 & 5061 & \\
\bottomrule
\end{tabular}
}
\label{tbl: min_pseudolabel_size}

\end{minipage}

\end{table}
Interestingly, applying intensity transform even hurts performance on liver. This implies that the configuration of intensity transforms in our experiments may deviate from the actual intensity distribution of livers in the dataset. 
\\
\noindent \textbf{Effect of pseudolabel sizes}:
To investigate the effect of pseudolabel sizes on performance, we experimented with pseudolabel sets with different minimum superpixel sizes. Table \ref{tbl: min_pseudolabel_size} shows that the granularity of superpixels should be reasonably smaller than sizes of actual semantic labels. This implies that too-coarse or too-fine-grained pseudolabels might divert the granularity of clusters in the learned representation space from that of real semantic classes. 
\section{Conclusion}
In this work, we propose a self-supervised few-shot segmentation framework for medical imaging. The proposed method successfully outperforms state-of-the-art methods without requiring any manual labeling for training. In addition, it demonstrates strong generalization to unseen semantic classes in our experiments. %This work potentially expands the future application of few-shot segmentation in medical images.
%Chen's suggestion:   
Moreover, the proposed superpixel-based self-supervision technique provides an effective way for image representation learning, opening up new possibilities for future works in semi-supervised and unsupervised image segmentation.
\\

\noindent\textbf{Acknowledgements.} 
This work is supported by the EPSRC Programme Grant EP/P001009/1. This work is also supported by the UK Research and Innovation London Medical Imaging and Artificial Intelligence Centre for Value Based Healthcare. The authors would like to thank Konstantinos Kamnitsas and Zeju Li for insightful comments.

\bibliographystyle{splncs}
%\bibliography{egbib}

%\input{./supps/supps.tex}
\end{document}